# On Advancements of the Forward-Forward Algorithm

Mauricio Ortiz Torres, Markus Lange, Arne P. Raulf

German Aerospace Center (DLR), Institut for AI-Safety and Security
Sankt Augustin and Ulm, Germany

*Abstract*— The Forward-Forward algorithm has evolved in machine learning research, tackling more complex tasks that mimic real-life applications. In recent years, it has been improved by several techniques to perform better than its original version, handling a challenging dataset like CIFAR10 without losing its flexibility and low memory usage. We have shown in our results that improvements are achieved through a combination of convolutional channel grouping, learning rate schedules, and independent block structures during training that lead to a 20% decrease in test error percentage. Additionally, to approach further implementations on low-capacity hardware projects, we have presented a series of lighter models that achieve low test error percentages within (21±3)% and a number of trainable parameters between 164,706 and 754,386. This serves as a basis for our future study on complete verification and validation of these kinds of neural networks.

## I. INTRODUCTION

Given the interest in low-power hardware in aerospace projects, we have that the Forward-Forward (FF) Algorithm [1], as introduced by its author Geoffrey Hinton in 2022, offers great potential for the realization of classification tasks of lower memory consumption. The FF algorithm employs two principles: First, the layers conforming to the network are trained independently by measuring their gradients with a locally defined "goodness" loss function, which measures the sum of the squares of the activity vectors in each layer. Second, the ground truth labels are overlaid on the training dataset to generate a subset of positive and negative datasets. The network is trained to maximize its goodness, favored by positive data and disfavored by negative data. The advantage of working with two forward passes shows itself during the training phase, because, in contrast to the traditional backpropagation, there is no need for storing neural activities or stopping to propagate error derivatives. This approach translates into lower memory requirements during training and also the inference phase. The algorithm has great flexibility for better hypertuning of its parameters and also for the realization of faster and more efficient inferences. Consequently, making it suitable for use in low-power hardware and possible parallelizations during training [2]. In this paper, we will present a summary of the latest research regarding this algorithm and its performance. Moreover, we will discuss the further improvements that have been made to the general structure of the algorithm, as well as our studies on lightweight FF models.

## II. THE FORWARD-FORWARD (FF) ALGORITHM

### A. Original approach

The architecture of the Forward-Forward algorithm is structured as a traditional neural network. However, its training procedure differs significantly, and it has two input spaces. These input spaces are given by positive and negative samples, which are created from the considered dataset of interest. Their creation for example for MNIST [3] data is as follows: Let $Y = \{y_1, \ldots, y_{10}\}$ be the set of labels, i.e., the different numbers in the MNIST dataset. Take an arbitrary image and its respective label vector $y_i$. To construct positive data samples, we one-hot encode $y_i$ and replace the first pixels of the image with the generated one-hot vector. To obtain a negative data sample from the chosen image, we randomly pick a label that is not $y_i$, e.g., $y_k \in Y$, $i \neq k$, and one-hot encode it. We then replace the first pixels of the image with the generated one-hot vector of the wrong label $y_k$. Thus, positive samples are overlaid with the correct label and negative samples with an incorrect one. In contrast to the usual backpropagation training, the FF algorithm operates with two forward passes of the network. One *positive* pass to adjust the weights of the hidden layer, $a^l$ [1], such that the goodness of each layer[2], $g^l(x) = \sum_i (a^l(x))_i^2$, increases for positive samples above a defined threshold value $\theta$ and one *negative* pass to decrease the goodness of the negative samples below the value of $\theta$. During a pass through the network, the output of each hidden layer is normalized to only transmit the relative information and not the explicit "goodness"-value from layer to layer.

The overall loss is

$$C_{\text{FF}} = C_{pos}^l + C_{neg}^l =$$
$$\ln[1 + \exp(\theta - g^l(x_{pos}))] + \ln[1 + \exp(g^l(x_{neg}) - \theta)]. \quad (1)$$

After the entire model has been trained under the "goodness" loss (1), the inference phase can be initiated by two different methods, referred to as one-pass inference and multi-pass inference. The "one-pass" inference works by training a unique SoftMax/Sigmoid layer with a general multi-classification loss function, such as the cross-entropy loss. The input is constructed by the activity vectors of each of the layers conforming the original network (Fig. 1). In particular, the activity vectors from the second[3] to the last layer of the network are concatenated into a new vector that is later on used as input for the SoftMax layer. The "multi-pass" inference works with the structure of the trained network but with multiple generated overlaid images as input. We employ the same procedure from the generation of positive and negative data to create N possible overlaid images for each of the N available labels. The overlaid images are then used one at a time in the forward pass, and the associated goodness values for each layer are collected and added up. By sampling all the possible overlaid images, we can collect the goodness for each layer and, with the use of an argmax function, determine the final label of the image (Fig. 2).

---

[1]Note that the output of a hidden layer is sometimes referred to as the activity vector of that layer.

[2]The evaluation of the goodness loss function for positive/negative samples is sometimes denoted as $g^l(x_{pos/neg}) \equiv g_{pos/neg}^l$.

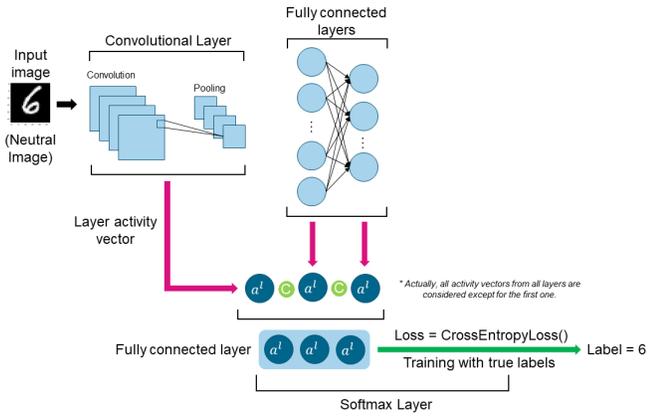

Fig. 1. Flow diagram for the one-pass inference step.

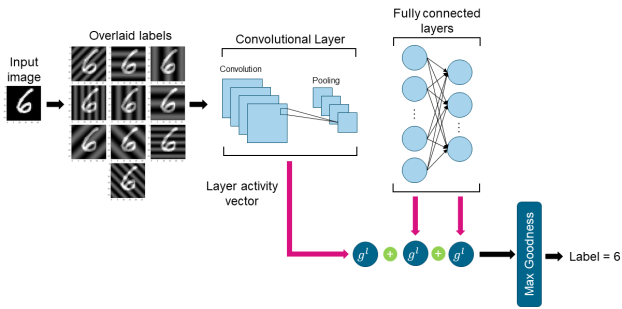

Fig. 2. Flow diagram for the multi-pass inference step.

### B. Variations to the algorithm

Along with some improvements to the model topology while using the original FF algorithm, there are several other approaches that try to deal with some of its weaknesses. We classify them according to the components that built the algorithm, starting from the creation of better correlated input data, improved training loss functions, more effective training routines, and faster inference schemes.

---

[3] It was shown that using the first hidden layer as part of input in the linear classifier leads to worse performance in the predictions [1].

*a) Creation of input data:* According to Hinton, the creation of negative input data must be done such that they have different long-range correlations but very similar short-range correlations[4]. However, achieving this can be more complicated when working with convolutional neural networks (CNNs). This is because the label information must be present over the entire image, ensuring that the multiple filters applied in each layer can capture features of the original image as well as the information on the classification space. One way of creating such data is to use learnable embeddings as the first layer of the network, as suggested in studies by Wu et al. [2] and Dooms et al. [4]. Alternatively, more elaborate procedures can be employed, such as the spatially extended labeling technique proposed by Scodellaro [5]. The technique introduces a superposition of the training dataset image with a second image of 2D Fourier modes of the same size as the training image. The mapping between labels and the specific characteristics of the Fourier modes (wavelength, amplitude, and orientation) can be freely chosen, allowing for the creation of multiple unique waves corresponding to each label in the classification problem.

Another approach is presented by Xing Chen et al. [6], where the creation of positive and negative samples is based on the principles of Noise Contrastive Estimation. Here, the input image is concatenated in two ways, either with itself or with an incorrect class image, creating thus a pair of positive+positive and positive+negative samples. This way of creating negative data can be treated as "noise data" and results in a distribution of negative examples that maintain a strong contrastive signal which later on enhances discriminative representations for learning.

One final approach consists of the substitution of positive and negative samples with convolutional group channels [7]. Here the output of each convolutional l-layer used in the network $Y^l \in \mathbb{R}^{N \times C \times H \times W}$[5] is subdivided into group input channels $\hat{Y}_j^l \in \mathbb{R}^{N \times S \times H \times W}$, where $S = C/J$ and $J$ corresponds to the numbers of classes present in the classification problem. This approach allows each convolutional layer to act as an independent classifier by having different goodness scores for the $J$ classes in the problem. In this construction, a holistic goodness factor for each layer is defined as

$$G_{n,j}^l = \frac{1}{S \times H \times W} \sum_{s=1}^{S} \sum_{h=1}^{H} \sum_{w=1}^{W} (\hat{Y}_{n,j,s,h,w}^l)^2 \quad (2)$$

and subsequently an associated positive and negative goodness is created for each convolutional layer, having thus $\mathbf{g}_{pos}^l = \mathbf{G}^l \cdot \mathbf{Z}^T \in \mathbb{R}^N$ and $\mathbf{g}_{neg}^l = \mathbf{G}^l \cdot (\mathbf{1} - \mathbf{Z}^T)$, where $\mathbf{Z} \in \{0,1\}^{N \times J}$ is the one-hot encoded vector of the true labels.

*b) Loss functions:* Variations of the loss function (1) have been considered under the scope of distance metric learning, where the construction of task-specific distance spaces gives better control over the distances from the data samples and their respective classes. New loss functions directly measure the discrepancy between projections of input patterns and labels. One such measure is given by the Symba loss function [4], $\sum_n \log[1+\exp((g_{pos}^l)_n-(g_{neg}^l)_n)]$. The direct difference between positive and negative goodness values reinforces the discrepancies between highly correlated positive and negative samples. Furthermore, Wu et al. proposes a loss function [2], $\max(m + (g_{pos}^l)_n - (g_{neg}^l)_n, 0) + \lambda(g_{neg}^l)_n$, that introduces a mixture between absolute and relative distances from the goodness measures in terms of a margin $m$ and regularization $\lambda$ variables. In the case of convolutional channel-wise competitive learning, the team from Andreas Papachristodoulou [7] introduces a channel-wise loss function, defined as,

$$C^l = -\frac{1}{N} \sum_{n=1}^{N} \log \left( \frac{\exp((g_{pos}^l)_n)}{\sum_{j=1}^{J} \exp(G_{n,j}^l)} \right). \quad (3)$$

---

[4] In the case of classical multi-layer perceptron (MLP), this can be easily achieved with the procedure mentioned in Section II-A.

[5] We work with the NCHW format, where $N$: Number of data samples in the batch, $C$: Image channels, $H$: Image height, $W$: Image width.

This method supports competitive learning through the channel dimension by calculating the probability of the target class over the total goodness score for each sample. Hence, this function works as a traditional cross entropy loss with goodness scores. Normalizing the goodness scores further supports a competitive class dynamic by balancing the scores and encouraging the increase of value for correct predictions and a decrease for the incorrect classes.

*c) Training routines:* Research on parallel training in neural networks, as conducted by Laskin et al. [8] and Xiong et al. [9], has led to explorations on architectures that benefit from allowing error signals to propagate through a neural network during training. By studying different training architectures on layer-wise trained models, it has been shown that these architectures can lead to an asynchronous layer-wise parallelism with a low memory footprint. The training architectures can be split into four types: 1) the traditional backpropagation, 2) the greedy local update[6], 3) the overlapping local update, and 4) the chunked local update. In these architectures, each group of layers is trained independently and learned through an auxiliary loss function, which updates the weights within the group. On the overlapping local update, the network is trained under a series of overlapped n-layers from the original network and composed of n-subsequent layers, in which every new group takes the last layer from the previous group as its initial layer. The chunked local update proceeds in building several chunks of n-layers sequentially ordered through the entire network without overlapping with other layers.

This way of training outperforms the original individual layer-wise training by having continue communication of the learned features in between layers. This approach has been shown to result in quick improvements on the learned features at each layer, as demonstrated by Dooms et al. [4], who found that this method can lead to significant enhancements in feature learning. Intuitively, there is a cooperation in the layers building the block to either enhance the local understanding of data or focus to increase the representations usefulness for the next layer. Furthermore, Wu et al. [2] propose an alternative approach, suggesting that random direct feedback connections can be integrated into the blocks to replace traditional back-propagation, allowing for more efficient weight updates using the calculated loss functions.

*d) Faster inference:* Using all activity vectors of a trained neural network might, for obvious reasons, be inefficient. Improved algorithms therefore directly employ the predictions estimated from each layer or a group of layers building the network. In the original paper, the model uses the activity vectors from the second-to-last layer. Dooms et al. [4] have shown that in more complex neural networks, the initial layers can be either detrimental or unnecessary for making accurate predictions, highlighting the importance of selecting the most relevant layers for analysis. The best results are thus obtained by considering the final layer or the last three layers of the model. Other studies like the one from Aminifar et al. [10] propose an approach to achieve lightweight inference by introducing a confidence variable in each layer, which allows for the selective inclusion of activity vectors and results in significant reductions in inference time.

Even though the original purpose of the algorithm focuses on its use in classification problems, the range of applicability of the algorithm or its underlying principles is investigated for unsupervised learning as well. In one study, Hwang et al. [11] introduce the Unsupervised learning Forward-Forward models (UFF), which train with the usual loss functions and without special inputs. However, the models are built up with cells instead of layers and are based on different unsupervised deep learning models, ranging from an auto-encoder- to a generative adversarial network-cell. Another study from Kumar et al. [12] build upon the FF algorithm to develop a novel variance-capturing auto-encoder, that allows to efficiently update data-driven models in real-time. Highlighting the model's capacity to learn fault detection and isolation in industry processes.

### III. RESULTS & ANALYSIS

We have developed an improved algorithm that has the following characteristics:

1) Creation of convolutional group channels for positive/negative sampling (Figure 3).
2) Use of the channel-wise loss function (3).
3) Training through chunked local updates that alternate in between iterations.
4) Realization of inference based on the activity vectors from the last two layers of the model.

Given the original and our improved algorithm, the different features explained in Section II-A and II-B are explored with a series of tests that give a particular focus on the plausible constructions of the network for the MNIST and CIFAR10 [13] datasets. While working with the original algorithm we employed for the MNIST dataset, a topology of $784 \times 500 \times 100$ for the MLP; and two convolutional layers of 32 and 64 filters plus a linear layer of $256 \times 100$ for the CNN. For CIFAR10, we considered a topology of $3072 \times 3072 \times 2000 \times 1000$ for the MLP, and four convolutional layers of 128, 264, 512, 1024 filters for the CNN. Our results in Table I show that the algorithm works correctly as a classifier and has similar results when employing either of the two inference options. One drawback of the one-pass inference is the additional softmax training time required to obtain the same order of error obtained from the multi-pass inference. Nevertheless, our tests have shown that the traditional one-pass inference results in less error percentage and faster inference times.

---
[6]Which in this case corresponds to the FF original training procedure.

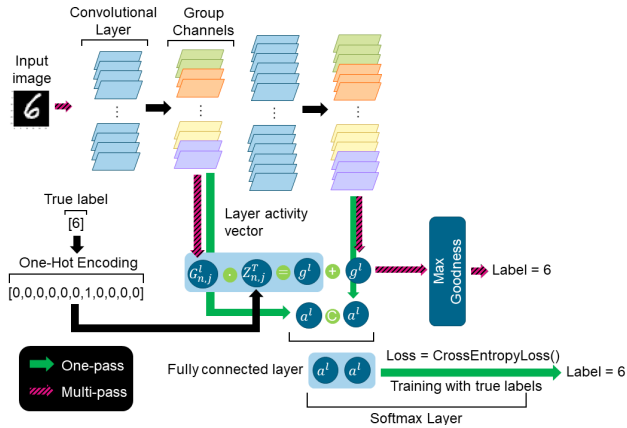

Fig. 3. Flow diagram for the inference step in our improved algorithm. The black arrows emphasize the creation of group channels and the use of the one-hot encoded matrix **Z** for the calculation of holistic goodness $\mathbf{G}^l$.

By adopting the improvements discussed in Section II-B, we repeated the same tests and obtained the results shown in Table II. In the improvements, we have employed a more elaborate network topology [4] which was trained using 300 epochs, where each layer consists of a batch normalization layer, followed by the convolution layer, a ReLU activation function, and an optional maxpool layer. Hence, using six convolutional layers of 128, 264 to 512 filters, with three maxpool applications. This model will be denominated by the name *FF_deep*.

TABLE I
TEST RESULTS WITH THE FF ALGORITHM USING THE MNIST AND CIFAR10 DATASETS.

| Dataset | Layers | Inference | Training Error [%] | Test Error [%] | Inference Time [s] |
|---|---|---|---|---|---|
| MNIST | MLP | One-pass | 8.6 | 8.0 | 0.6 |
|  | MLP | Multi-pass | 7.3 | 7.2 | 2.2 |
|  | CNN | One-pass | 4.5 | 4.4 | 0.1 |
|  | CNN | Multi-pass | 4.9 | 5.0 | 1.9 |
| CIFAR10 | MLP | One-pass | 39.2 | 39.3 | 0.1 |
|  | MLP | Multi-pass | 41.1 | 41.2 | 0.2 |
|  | CNN | One-pass | 20.5 | 44.3 | 1.4 |
|  | CNN | Multi-pass | 44.8 | 48.3 | 4.5 |

The results show a considerable test error percentage reduction by more than 20% compared to the original algorithm. A multi-step learning rate schedule was necessary for better stability and convergence of the results. From all the improved techniques, we want to highlight the convolutional channel grouping [7] since this approach increases the network's capacity to learn intraclass features. This method alone is responsible for allowing the model to obtain 27% in test errors. Additionally, the channel-wise loss enforces a way of training mutually exclusive filters that contribute to the most important features needed for each class. Later on, with the depth of the network, these features attain more complex representations and build up a unique set of filters for each class that are not directly shared with other classes.

TABLE II
TEST RESULTS WITH THE IMPROVED FF ALGORITHM USING THE MNIST AND CIFAR10 DATASETS.

| Dataset | Layers | Inference | Training Error [%] | Test Error [%] | Inference Time [s] |
|---|---|---|---|---|---|
| MNIST | CNN | One-pass | 0.2 | 0.8 | 1.7 |
|  | CNN | Multi-pass | 0.2 | 0.7 | 1.4 |
| CIFAR10 | CNN | One-pass | 1.2 | 18.8 | 2.1 |
|  | CNN | Multi-pass | 1.8 | 18.2 | 1.8 |

The improved approach enables faster inference times for the multi-pass scheme, however, it is still subject to greater inference times depending on the number of activation vectors. This characteristic is avoided by the one-pass inference step because it's only dependent on the final trained model that initially considered the number of activity vectors before training. Given that the improved algorithm has an independent block structure during training, the benefits from efficient hyperparameter tuning of the models are still present in this case. Also, the features extracted at previous layers by the channel-wise competitive learning are efficiently passed to further layers to enhance the representations of the entire network. Hence continuously increasing the accuracy of all the network's layers.

The proposed algorithm presents some specific characteristics that make it very suitable for low-capacity hardware projects. Firstly, avoiding backpropagation allows for fewer gradients to be computed during training, and all activation vectors do not need to be kept in memory for the backward pass. This has the advantage that the model can learn while pipelining sequential data through the network without saving the activity vectors of intermediate layers or propagating error derivatives. The independent training of each layer or block presents better regulation on the hyperparameters of the model and thus results generally in better convergence of its gradients. This also prevents further propagation of vanishing or exploding gradients to subsequent layers. The algorithm reduces the computational process by requiring fewer activity vectors during inference, especially by the multi-pass inference scheme, where the new algorithm eliminates the need to create overlaid images. Finally, accelerations on parallel computing architectures can be achieved due to the training independence of each layer.

To tackle further implementations on low-capacity hardware projects, we have explored the possibilities of lighter FF models based only on CNN layers with the mentioned configuration of techniques. The goal was to achieve a noticeable reduction in parameters, because the *FF_deep* model is based on an architecture in [4] where their shallow network has trained 4,100,000 parameters for the last two activation vectors in the inference phase. In Table III we show a list of the studied light FF models along with the *FF_deep* model and their respective architectures[7]. Similarly, in Table IV the results on all the light FF models are provided. The models have a test error of (21±3)% and consider a different range of trainable parameters. Our smallest network, *FF_tiny*, has 96% less parameters compared to *FF_deep*, but is subject to the highest test error of 24.1%. In contrast, *FF_optimal* conserves a test error below 20% and a parameter reduction of 81% in comparison to our largest model.

TABLE III
LIGHT FF MODELS ARCHITECTURES TRAINED WITH THE IMPROVED FF ALGORITHM.

| Model name | Channels; Kernel Size | Trainable Parameters |
|---|---|---|
| *FF_tiny* | [3, 50, 50, 50, 50]; [3, 3, 3, 4] | 164,706 |
| *FF_small* | [3, 50, 50, 70, 70]; [3, 3, 3, 4] | 239,266 |
| *FF_medium* | [3, 50, 50, 100, 150]; [3, 3, 3, 4] | 484,506 |
| *FF_optimal* | [3, 50, 50, 100, 160, 160]; [3, 3, 3, 4, 3] | 754,386 |
| *FF_deep* | [3, 130, 130, 260, 260, 260, 510]; [3, 3, 3, 5, 3, 3] | 4,131,346 |

TABLE IV
LIGHT FF MODELS TEST RESULTS WITH THE IMPROVED FF ALGORITHM USING THE CIFAR10 DATASET.

| Model name | Inference | Training Error [%] | Test Error [%] |
|---|---|---|---|
| *FF_tiny* | One-pass | 15.3 | 24.4 |
| | Multi-pass | 17.9 | 24.1 |
| *FF_small* | One-pass | 12.3 | 23.4 |
| | Multi-pass | 14.8 | 23.1 |
| *FF_medium* | One-pass | 8.0 | 21.5 |
| | Multi-pass | 10.1 | 20.7 |
| *FF_optimal* | One-pass | 2.0 | 19.6 |
| | Multi-pass | 2.4 | 19.3 |
| *FF_deep* | One-pass | 1.2 | 18.8 |
| | Multi-pass | 1.8 | 18.2 |

During training, the models are trained using 50-70 epochs. Such a small value shows that the training methods employed lead to quick convergence of the gradients. Nevertheless, a careful use of the MultiStep learning schedule with 3-5 milestones in between the epochs and $\gamma \in \{0.1, 0.2\}$[8], pushes the models to the optimal 18% test error. The training process was susceptible to unstable gradients that propagate through the whole network, given our training technique, because of our choice to obtain shallow networks. This set of light FF models offers broad flexibility to realize further analysis on low-power hardware, without giving up accuracy on the models. In particular, we are employing this flexibility in an upcoming work on the verification and validation of FF models.

For the case of the *FF_optimal* model, we balanced the trade-off between parameter reduction and test error by considering three characteristics. First, using a tested minimum number of 160 filters at the final convolutional layer to generate test errors under 20% for all runs. Second, adding the same number of filters in the final layer enabled the model to stabilize the results and permits efficient reuse of learned features to achieve richer representations. Lastly, employing four milestones on the learning rate schedules further stabilized the gradients without decreasing the individual layer accuracy.

It is worth mentioning that further compression techniques can be utilized with this algorithm, like the case of quantization schemes. We have considered this idea for the case of the *FF_deep* model by using the quantization framework Brevitas from Xilinx.[9] The provided tools are well adapted for our use case, since they offer a series of individual quantize layers that behave analogously to those in traditional neural networks. We were able to construct the same model in their quantized version and carry out quantization-aware training. This resulted in a test error of 22.3% for layers with 2 bit-widths. Further exploration in this area is needed, but the initial results show that the procedure is indeed possible.

IV. CONCLUSION

The research on the Forward-Forward algorithm keeps evolving into further areas of machine learning and pursues more demanding tasks close to real-life applications. The algorithm continues to show its great potential for usage in low-power hardware, given its capabilities of layer- or block-wise training, and its simplified computational process. In our results, we have seen that the improved algorithm performs better than the original one by showing an approximate 20% decrease in test error percentage. Moreover, it can handle more challenging datasets without losing its flexibility.

The improvements in the generation of input data and training routines open the possibility for different learning configurations that allow a faster and more efficient flow of information between the layers. However, the group architectures must be limited since otherwise they could fall back into traditional backpropagation and lose the benefits from the FF architecture. The presented lighter FF models show how this configuration of techniques leads to a broad range of smaller models that still achieve low test error percentages within (21±3)% and a number of trainable parameters between 164,706 and 754,386. Hence, these models have a high performance comparable to those analyzed in state-of-the-art FF papers while keeping a considerably smaller size. As a next step, we will investigate the benefits provided by these kinds of models to achieve complete verification and validation of the neural network.

---

[7]The underline numbers in the kernel size denote that the maxpool operation was also used at that same location of the network.

[8]For the MultiStepLR class defined in PyTorch, $\gamma$ is associated with the multiplicative factor of learning rate decay.

[9]The framework is available in the Github: https://github.com/Xilinx/brevitas